\PassOptionsToPackage{table}{xcolor}
\AddToHook{package/cleveref/before}{%
  \RequirePackage{amsmath,amssymb,amsthm}%
}
\documentclass{bytedance}

\usepackage{amsmath,amsfonts,bm}

\def\eqref#1{equation~\ref{#1}}

\def\1{\bm{1}}

\DeclareMathAlphabet{\mathsfit}{\encodingdefault}{\sfdefault}{m}{sl}
\SetMathAlphabet{\mathsfit}{bold}{\encodingdefault}{\sfdefault}{bx}{n}

\usepackage{url}
\usepackage{xspace}
\usepackage{float}
\usepackage{wrapfig}
\usepackage{pifont}
\usepackage{algorithm}
\usepackage{algpseudocode}

\newcommand{\ourmethod}{DuoMatching\xspace}
\newcommand{\submethod}{LatentBridge\xspace}

\newcommand{\myparagraph}[1]{\vspace{0.1cm}\noindent\textbf{#1}}

\theoremstyle{plain}

\definecolor{oursgray}{gray}{0.94}

\title{\ourmethod: Joint-Marginal Distribution Matching for Few-Step Video Generation}

\author{%
  \centerline{%
    Jiahao Zhan$^{1,2}$\quad
    Yan Wang$^{2}$\quad
    Yongrui Ma$^{1,2}$\quad
    Qunliang Xing$^{2}$%
  }
  \centerline{%
    Ruchang Yao$^{1}$\quad
    Runtao Liu$^{3}$\quad
    Shijie Zhao$^{2\dagger}$\quad
    Tianfan Xue$^{1,4\dagger}$%
  }
}

\affiliation[]{%
  $^{1}$MMLab, CUHK \quad
  $^{2}$ByteDance Inc. \quad
  $^{3}$HKUST \quad
  $^{4}$CPII under InnoHK
}

\abstract{
Streaming video generation has benefited from distribution matching distillation (DMD), which matches the joint distribution of video frames to a video teacher’s approximation of the real video distribution.
Although this joint matching mitigates drift during autoregressive rollouts, limitations remain in visual quality and semantic alignment.
To address these limitations, we propose \ourmethod, a distribution matching framework that approximates the real video distribution through a unified joint-marginal formulation.
On top of existing joint matching formulations, the additional marginal matching objective provides dedicated frame-level supervision from an image generator, transferring complementary visual and semantic priors from it.
To apply this frame-level supervision in video generation, we introduce \submethod to resolve the latent representation mismatch between the video student and the image teacher.
Latent Variation Sampling further distributes such frame-level supervision across distinct temporal segments, reducing redundancy.
Experiments demonstrate that \ourmethod improves visual quality, composition, and semantic alignment while largely preserving motion dynamics. 
Human evaluations show overall preference rates above 80\% against all evaluated baselines.
}

\date{October 2, 2026}
\checkdata[Project Page]{\url{https://johnzhan2023.github.io/DuoMatching/}}

\hypersetup{
  pdftitle={DuoMatching: Joint-Marginal Distribution Matching for Few-Step Video Generation},
  pdfauthor={Jiahao Zhan, Yan Wang, Yongrui Ma, Qunliang Xing, Ruchang Yao, Runtao Liu, Shijie Zhao, Tianfan Xue},
  pdfsubject={Few-step video generation},
  pdfkeywords={video generation, distribution matching distillation, DuoMatching}
}

\begin{document}
\maketitle
\suppressfloats[t]

\begingroup
  \renewcommand{\thefootnote}{\fnsymbol{footnote}}
  \footnotetext[2]{Corresponding authors.}
\endgroup

\begin{figure}[t]
    \centering
    \includegraphics[width=\textwidth]{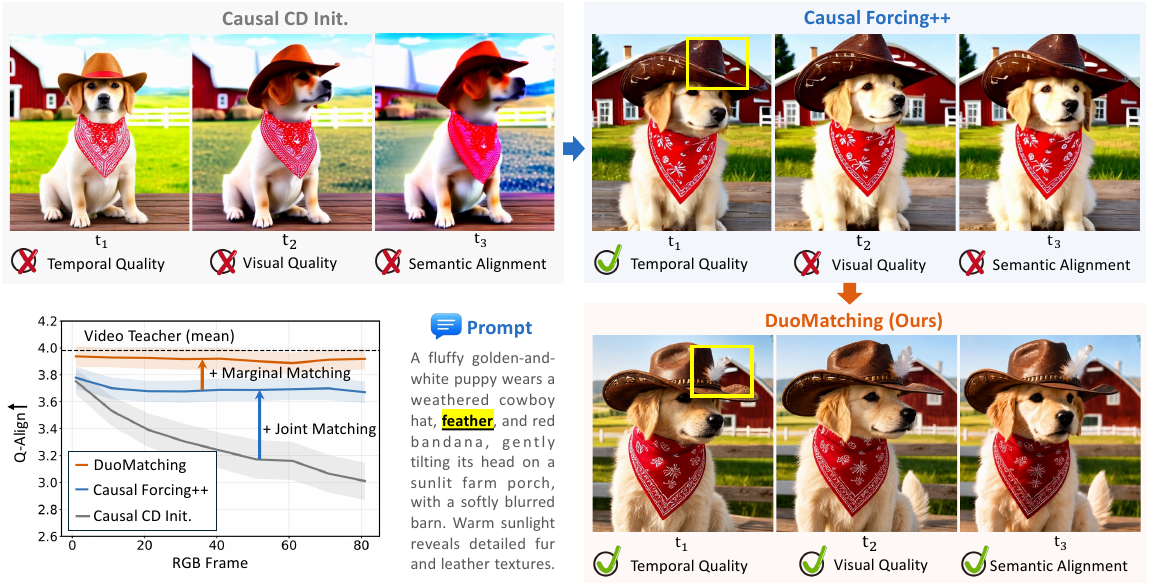}
\caption{
Motivation and results of \ourmethod.
Videos generated by autoregressive models often have degraded quality for later rollout frames (see Q-Align~\citep{wu2023q} curves).
Joint DMD mitigates this drift, yet the student still retains a visual quality gap relative to the video teacher.
The example reveals a missing feather specified in the prompt.
To solve this problem, we reformulate distribution matching from a coupled joint-marginal perspective to better approximate the real video distribution, as illustrated by improved visual quality and the recovered feather, while maintaining temporal coherence. 
}
    \label{fig:teaser}
\end{figure}

\section{Introduction}
\label{sec:intro}

Recent video generation models have achieved remarkable advances in visual quality and temporal coherence~\citep{liu2024sora,yang2025cogvideox}. 
Despite their impressive performance, these models typically require many sequential denoising steps, resulting in substantial computational cost and inference latency that hinder real-time applications such as interactive world simulation~\citep{parker2025genie} and digital entertainment~\citep{zhao2026sana}. 
Recently, distribution matching distillation (DMD)~\citep{yin2024one} addresses this computational bottleneck by distilling multi-step diffusion~\citep{ho2020denoising} or flow models~\citep{lipman2022flow} into few-step generators.
Although streaming video generation built on this paradigm has approached real-time inference~\citep{sun2025worldplay,yang2025longlive,liu2026realwonder}, preserving generation quality under distribution matching remains a challenge~\citep{feng2026one}.

Specifically, we identify remaining limitations in visual quality and semantic alignment of autoregressive generators trained with existing DMD methods.
These methods use the video teacher's distribution as a proxy for the real video distribution~\citep{yin2024one} and match the joint distribution of all frames, a formulation we refer to as joint DMD.
As shown in Figure~\ref{fig:teaser}, the quantitative results indicate that joint DMD substantially mitigates drift over the rollout, but yields only limited improvement in initial-frame quality.
The qualitative comparison highlights remaining limitations in visual quality and semantic alignment, as illustrated by insufficient fine-grained fur detail and the missing feather in this example.
These frame-level deficiencies suggest that existing DMD methods do not adequately model the marginal distribution, i.e., the distribution of individual frames.
Indeed, existing DMD matches marginals jointly with cross-frame dependencies, without a dedicated objective for individual frames.
Adding an additional marginal distribution constraint in training may better approximate the real video distribution.

\begin{wrapfigure}{r}{0.5\textwidth}
    \centering
    \vspace{-0.5cm}
    \includegraphics[width=\linewidth]{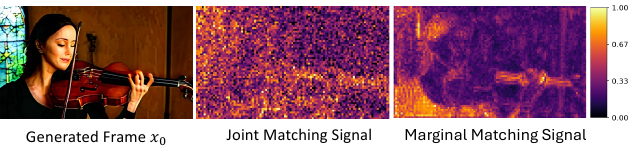}
    \par
\caption{
Spatial structure of score differences in joint and marginal distribution matching.
Marginal matching exhibits more spatially coherent responses, whereas joint matching produces more scattered responses in this example. Each heatmap is independently normalized to highlight spatial structure rather than absolute magnitude.
}
\vspace{-0.3cm}
    \label{fig:score_heatmap}
\end{wrapfigure}

To model the marginal distribution in video generation, we employ image generation models~\citep{seedream2025seedream,cai2025z} which provide a natural frame-level reference distribution.
Their strong priors support fine-grained appearance modeling, and their semantic alignment capabilities also help capture prompt-specified attributes and spatial relationships.
Marginal distribution matching directly steers the student's frame-level distribution toward the image teacher's distribution, without being constrained by cross-frame dependencies.
As illustrated in Figure~\ref{fig:score_heatmap}, the marginal distribution matching's signal exhibits more coherent spatial structure, with particularly strong responses on the subject's overly smooth black clothing.
Together with the video teacher's joint distribution, the additional marginal distribution from image teachers guides the student to generate more realistic video.

We propose \ourmethod, a distribution matching framework that approximates the real video distribution through a unified joint-marginal formulation.
For few-step video generation, we instantiate this formulation by training a video generator with a weighted combination of joint DMD and marginal DMD.
Specifically, we sample a subset of frames from each generated video and apply marginal DMD independently to each frame using the image teacher, transferring visual and semantic priors to strengthen frame-level supervision. 
Joint DMD supervises the full sequence using the video teacher, constraining cross-frame dependencies to encourage temporal consistency.

Still, applying marginal distribution matching to a video student is nontrivial.
Applying marginal DMD requires frame-level representations compatible with the image teacher. 
However, temporally compressed video latents do not directly correspond to individual frames, and video and image VAEs may use incompatible latent spaces.
Therefore, we propose \textbf{\submethod}, a lightweight module that resolves the latent representation mismatch, enabling frame-level distribution matching without substantially suppressing encoded dynamics. 
We further introduce \textbf{Latent Variation Sampling} (LVS) to distribute marginal matching across distinct temporal segments, reducing redundant supervision to better balance visual quality and video dynamics.

Experiments demonstrate that \ourmethod improves fine-grained visual quality, visual composition, and semantic alignment not only in causal generation but also in bidirectional generation, while largely preserving temporal quality.
Human evaluations further show overall preference rates above 80\% against all evaluated baselines. 

\section{Related Work}
\myparagraph{Few-step video generation.}
Recent video generation models~\citep{liu2024sora,seedance2026seedance,veo3} achieve strong visual quality but rely on expensive iterative denoising, motivating the development of few-step generation methods. Some works adopt adversarial post-training to enable one-step video generation~\citep{lin2025diffusion}, while others reduce sampling steps through model distillation~\citep{li2024t2v1, ding2025dollar}. VideoLCM~\citep{wang2023videolcm} extends consistency distillation to video generation. 
CausVid~\citep{yin2025slow} uses DMD~\citep{yin2024improved} for autoregressive video generation, enabling KV-cache reuse. 
Subsequent works address different limitations of causal distillation: Self Forcing~\citep{huang2026self} reduces the mismatch between training and self-generated inference rollouts, while Causal Forcing~\citep{zhu2026causal} addresses the architectural mismatch to ensure frame-level injectivity. \ourmethod instead investigates the persistent quality limitations of distribution matching itself to improve the generation quality of few-step generators.

\myparagraph{Distribution matching.}
Distribution matching provides a general framework for generative modeling by aligning the generated distribution with the real data distribution~\citep{yin2024improved,yin2024one, wang2023prolificdreamer}. However, the matching process can still suffer from optimization instability. Phased DMD~\citep{fan2026phased} eases multi-step optimization by decomposing the denoising process into smaller SNR intervals. Beyond improving the optimization itself, several works introduce additional supervision to enhance the matched distribution. One-Forcing~\citep{feng2026one} adds adversarial supervision to alleviate frame blurring at extremely low NFE, while Reward Forcing~\citep{lu2026reward} and DMDR~\citep{jiang2025distribution} incorporate reward signals to favor preferred outputs. Data-Forcing~\citep{chen2026data} leverages real data to mitigate the mode-seeking behavior of DMD, whereas SA-DMD~\citep{xiao2026joyai} anchors distribution matching to the source video for better content preservation in video editing. 
\ourmethod complements joint distribution matching with direct marginal distribution matching to more stably approximate the real video distribution.

\myparagraph{Image priors for video generation.}
Image priors support video generation through image animation~\citep{tian2021good,guo2024i4vgen}, and initialization from pretrained image models~\citep{singer2022make,blattmann2023stable,bai2026instructionbasedvideoeditingrepurposing}. 
Beyond initialization, image generation models have been integrated into video sampling to improve visual quality~\citep{zhang2025videoelevator,shao2025iv}.
Some works have also explored incorporating image priors through distribution matching to accelerate video generation, focusing on image-derived video models such as AnimateDiff~\citep{guo2023animatediff}.
MCM~\citep{zhai2024motion} introduces a discriminator trained on image data during consistency distillation, while AVDM2~\citep{zhu2024accelerating} incorporates score distribution matching using an image teacher to enhance video adversarial training.
We revisit distribution matching from a joint-marginal perspective, theoretically analyze the relationship between joint and marginal matching, and formulate both as distribution matching objectives within \ourmethod.
We further introduce \submethod to generalize this framework to modern video generators with temporally compressed latents.

\section{Method}
\begin{figure*}[t]
    \centering
    \includegraphics[width=1.0\textwidth]{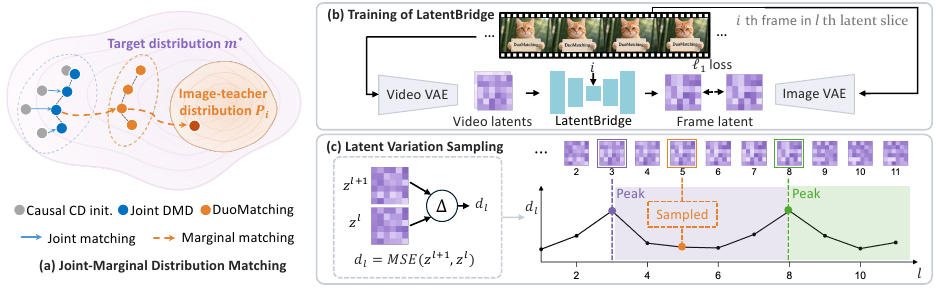}
\caption{
    \textbf{Overview of \ourmethod.}
    (a) \ourmethod complements joint distribution matching with an image-teacher reference distribution $P_i$. Each group schematically represents a video, with nodes depicting frame-level distributions.
    (b) \submethod is trained to map a video latent $z^l$, conditioned on the preceding slice $z^{l-1}$ and a local frame index $i$, to the corresponding image latent using an $\ell_1$ reconstruction loss.
    (c) Latent Variation Sampling uses differences between adjacent latent slices to partition the sequence into segments, within which slices are randomly sampled.
}
    \label{fig:method}
\end{figure*}

We first formulate video generation as distribution matching, identify the limitations of relying solely on joint matching, and motivate explicit frame-level marginal matching in Section~\ref{sec:distribution_matching}.
In Section~\ref{sec:joint_marginal}, we reformulate distribution matching through a unified joint-marginal framework to better approximate the real video distribution.
To enable marginal distribution matching with image generation models, we introduce \submethod to resolve the mismatch between video and image latent spaces in Section~\ref{sec:latent_bridge}.
Finally, Section~\ref{sec:mc_sampling} introduces Latent Variation Sampling (LVS) to improve temporal coverage and reduce redundancy in marginal matching.

\subsection{Distribution Matching for Video Generation}
\label{sec:distribution_matching}
Video generation aims to learn a generator $G_\theta$ whose output distribution $Q_\theta$ matches the real data distribution $P^\star$. For simplicity, we omit text conditioning from the notation. The ideal distribution matching objective is
\begin{equation}
\mathcal{J}_{\mathrm{DM}}(\theta)
=
D_{\mathrm{KL}}\!\left(Q_\theta \,\|\, P^\star\right).
\label{eq:ideal_distribution_matching}
\end{equation}
Since the score of $P^\star$ is not directly available, we follow prior work and approximate it using a pretrained video teacher with distribution $P_v$, which leads to the joint distribution matching objective:
\begin{equation}
\mathcal{J}_{\mathrm{joint}}
=
D_{\mathrm{KL}}\!\left(Q_\theta \,\|\, P_v\right).
\label{eq:video_dmd}
\end{equation}
Joint distribution matching is performed over all temporal latent slices. In practice of joint DMD, given a generated video latent $\mathbf{z}=G_\theta(\boldsymbol{\xi})$, its noisy version is $\mathbf{z}_\tau=\alpha_\tau\mathbf{z}+\sigma_\tau\boldsymbol{\epsilon}$, where $\boldsymbol{\xi}$ and $\boldsymbol{\epsilon}$ are independent standard Gaussian noise. 
The video teacher and an auxiliary score model then estimate the teacher and student scores at $\mathbf{z}_\tau$, respectively, to guide the generator update.
We write $\mathbf{z}_\tau=(z_\tau^1,\ldots,z_\tau^L)$ and use $\mathbf{z}_\tau^{\setminus l}$ to denote all slices except $z_\tau^l$. For either the teacher or student density $p_\tau$, the joint distribution factorization yields
\begin{equation}
\nabla_{z_\tau^l}\log p_\tau(\mathbf{z}_\tau)
=
\nabla_{z_\tau^l}\log p_\tau\!\left(z_\tau^l\mid\mathbf{z}_\tau^{\setminus l}\right).
\end{equation}
Thus, the score for each latent slice is conditioned on the other slices. 
Although joint distribution matching also constrains frame-level marginals, a change that improves an individual frame may not be favored if it conflicts with the surrounding frames. 
For example, adding a missing feather or finer fur details to one frame alone could introduce such inconsistencies with other frames. 
This potential tension, together with the observed quality limitations in Figure~\ref{fig:teaser}, motivates an additional marginal distribution objective.

\subsection{\ourmethod: Joint-Marginal Distribution Matching}
\label{sec:joint_marginal}
Let $M$ denote a frame-sampling operator that selects one frame from a video according to a specified sampling policy. 
The corresponding frame-level marginals of the video teacher and real data are $m_v=MP_v$ and $m^*=MP^*$, respectively.
Since $m^*$ is defined over individual frames, it can also be directly approximated by an image generation model with distribution $P_i$. 
Under ideal marginal fitting within the same feasible family, the $D_{\mathrm{KL}}(m^*\|P_i)$ is no greater than the $D_{\mathrm{KL}}(m^*\|m_v)$ (see Appendix~\ref{app:marginal_reference} for the assumptions and proof).
Therefore, we extend the original distribution matching objective from a single proxy distribution to joint and marginal references, forming our \ourmethod framework.

Let $m_\theta=MQ_\theta$ denote the generator's frame-level marginal. We retain the video teacher as the joint distribution reference and use the image teacher for direct marginal supervision, defining the joint-marginal distribution matching objective:
\begin{equation}
\mathcal J_{DM}(\theta)
=
D_{\mathrm{KL}}(Q_\theta\|P_v)
+\omega D_{\mathrm{KL}}(m_\theta\|P_i),
\qquad \omega>0,
\end{equation}
where $\omega$ balances the influence of the video teacher's joint distribution estimate and the image teacher's complementary marginal estimate. 
Under the assumptions detailed in Appendix~\ref{app:joint_marginal_improvement}, we show that there exists $\omega>0$ for which the ideal joint-marginal optimum is closer to $P^*$ than $P_v$ in forward KL.
As illustrated in Figure~\ref{fig:method}(a), the marginal term strengthens frame-level distribution matching beyond joint supervision alone, while transferring complementary visual and semantic priors from the image teacher.
The joint term simultaneously constrains cross-frame dependencies, encouraging the student to incorporate these priors coherently throughout the generated sequence.
Implementing the joint and marginal terms with their corresponding DMD surrogates gives:
\begin{equation}
\mathcal L_{\mathrm{DuoMatching}}
=
\mathcal L_{\mathrm{joint\text{-}DMD}}
+\omega\mathcal L_{\mathrm{marginal\text{-}DMD}}.
\end{equation}
In practice, joint DMD follows the procedure in Causal Forcing~\citep{zhu2026causal}. For marginal DMD, given a generated video latent $z=G_\theta(\xi,c)$, \submethod maps a temporal latent slice $z^l$ and a local frame index $i$ to an image-compatible representation $u^{l,(i)}$. The mapping is detailed in Section~\ref{sec:latent_bridge}. We perturb this representation according to the image teacher's noise schedule: $u_\tau^{l,(i)}=\alpha_\tau^{\mathrm{img}}u^{l,(i)}+\sigma_\tau^{\mathrm{img}}\epsilon$, where $\epsilon\sim\mathcal N(0,I)$.
A frozen image teacher and a trainable fake score estimator provide score estimates $s_{\mathrm{real}}^{\mathrm{img}}$ and $s_{\mathrm{fake}}^{\mathrm{img}}$ at $(u_\tau^{l,(i)},\tau,c)$, respectively. The fake estimator is trained on the mapped generated latents to track their evolving distribution. The marginal DMD gradient is estimated as
\begin{equation}
\nabla_\theta\mathcal L_{\mathrm{marginal}}
=
\mathbb E_{\xi,l,i,\tau,\epsilon}
\left[
w_{\mathrm{img}}(\tau)
\left(s_{\mathrm{fake}}^{\mathrm{img}}-s_{\mathrm{real}}^{\mathrm{img}}\right)^\top
\frac{\partial u_\tau^{l,(i)}}{\partial\theta}
\right],
\end{equation}
where $w_{\mathrm{img}}(\tau)$ is a timestep-dependent weighting function. The sampling of $l$ is described in Section~\ref{sec:mc_sampling}. Gradients propagate through the frozen LatentBridge to the student generator, providing direct frame-level supervision. Although motivated by the quality limitations in autoregressive generation, our joint-marginal framework also generalizes to bidirectional generators.

\subsection{\submethod: Bridging Video and Image Latents}
\label{sec:latent_bridge}

Because we use image generation models to approximate the marginal distribution, an additional issue arises from the difference between image VAE encoding and video VAE encoding.
Most video VAEs~\citep{zhao2024cv} use temporal compression, where one video latent maps to multiple actual RGB frames, while each image latent is constrained to one frame.
A straightforward solution is to decode video latents into RGB frames and re-encode them with the image VAE, but backpropagating matching gradients through both VAEs incurs substantial memory overhead. 

We therefore introduce \submethod, a lightweight differentiable module that maps temporally compressed video latents to frame-specific representations in the image teacher's latent space. We pretrain this module using paired video and image latents, as illustrated in Figure~\ref{fig:method}(b).

Given a temporal latent slice $\mathbf{z}^{l}$, we uniformly sample a local frame index $i\sim\mathcal{U}\{1,\ldots,r\}$, where $r$ is the number of RGB frames represented by that slice. To account for the causal temporal encoding of the video VAE, we additionally condition on the preceding latent slice $\mathbf{z}^{l-1}$, providing temporal context for the mapping. Conditioned on these slices and $i$, \submethod predicts the latent of the corresponding RGB frame in the image teacher's latent space:
\begin{equation}
    u^{l,(i)}=B_\phi(\mathbf{z}^{l-1},\mathbf{z}^{l},i).
    \label{eq:latent_bridge}
\end{equation}
For training, let $x^{l,i}$ denote the RGB frame corresponding to latent slice $l$ and local frame index $i$. The target latent is obtained by encoding $x^{l,i}$ with the image teacher's VAE encoder $E_{\mathrm{img}}$. We train $B_\phi$ using an $\ell_1$ reconstruction loss:
\begin{equation}
    \mathcal{L}_{\submethod}
    = \mathbb{E}_{\mathbf{z}^{l-1},\mathbf{z}^{l},i}
    \left[
        \left\|B_\phi(\mathbf{z}^{l-1},\mathbf{z}^{l},i)-E_{\mathrm{img}}(x^{l,i})\right\|_1
    \right].
    \label{eq:bridge_pretraining}
\end{equation}
By conditioning on the sampled frame index, \submethod learns to recover distinct frame-level representations from shared temporal latent slices. 
Optimization through \submethod can improve the quality of decoded video frames (see Appendix~\ref{app:latent_bridge_transfer} for an analysis of frame-level loss transfer).
The module can be trained for the latent space of a chosen image teacher, allowing \ourmethod to use different image priors.

\subsection{Latent Variation Sampling for Frame-Level Supervision}
\label{sec:mc_sampling}

Marginal distribution matching supervises sampled frames, but concentrating samples in slowly changing regions produces redundant supervision. 
This incurs unnecessary computational cost and may suppress motion dynamics by overemphasizing nearly static content.
To address the issue, we introduce Latent Variation Sampling, as shown in Figure~\ref{fig:method}(c). Given the clean video latent sequence $\mathbf{z}=(\mathbf{z}^{1},\ldots,\mathbf{z}^{L})$, we compute the mean squared difference between adjacent latent slices:
\begin{equation}
    d_l=\frac{1}{CHW}\left\|\mathbf{z}^{l+1}-\mathbf{z}^{l}\right\|_F^2,
    \qquad l=1,\ldots,L-1,
    \label{eq:latent_variation}
\end{equation}
where $C$, $H$, and $W$ are the channel and spatial dimensions of each latent slice. We select the $K-1$ positions with the largest MSE values and split the sequence into $K$ contiguous temporal segments. Denoting the sorted split positions by $b_1<\cdots<b_{K-1}$, with $b_0=0$ and $b_K=L$, we uniformly sample one latent slice from each segment:
\begin{equation}
    l_k\sim\mathcal{U}\{b_{k-1}+1,\ldots,b_k\},
    \qquad k=1,\ldots,K.
    \label{eq:lvs_sampling}
\end{equation}
For each selected slice $\mathbf{z}^{l_k}$, we apply marginal DMD to the image-aligned latent $u^{l_k,(i)}$, with $i$ uniformly sampled. Each segment receives one frame sample per update, distributing the limited supervision budget across temporal regions. 

\section{Experiments}
\begin{figure}[!t]
    \centering
    \includegraphics[width=\textwidth,height=0.70\textheight,keepaspectratio]{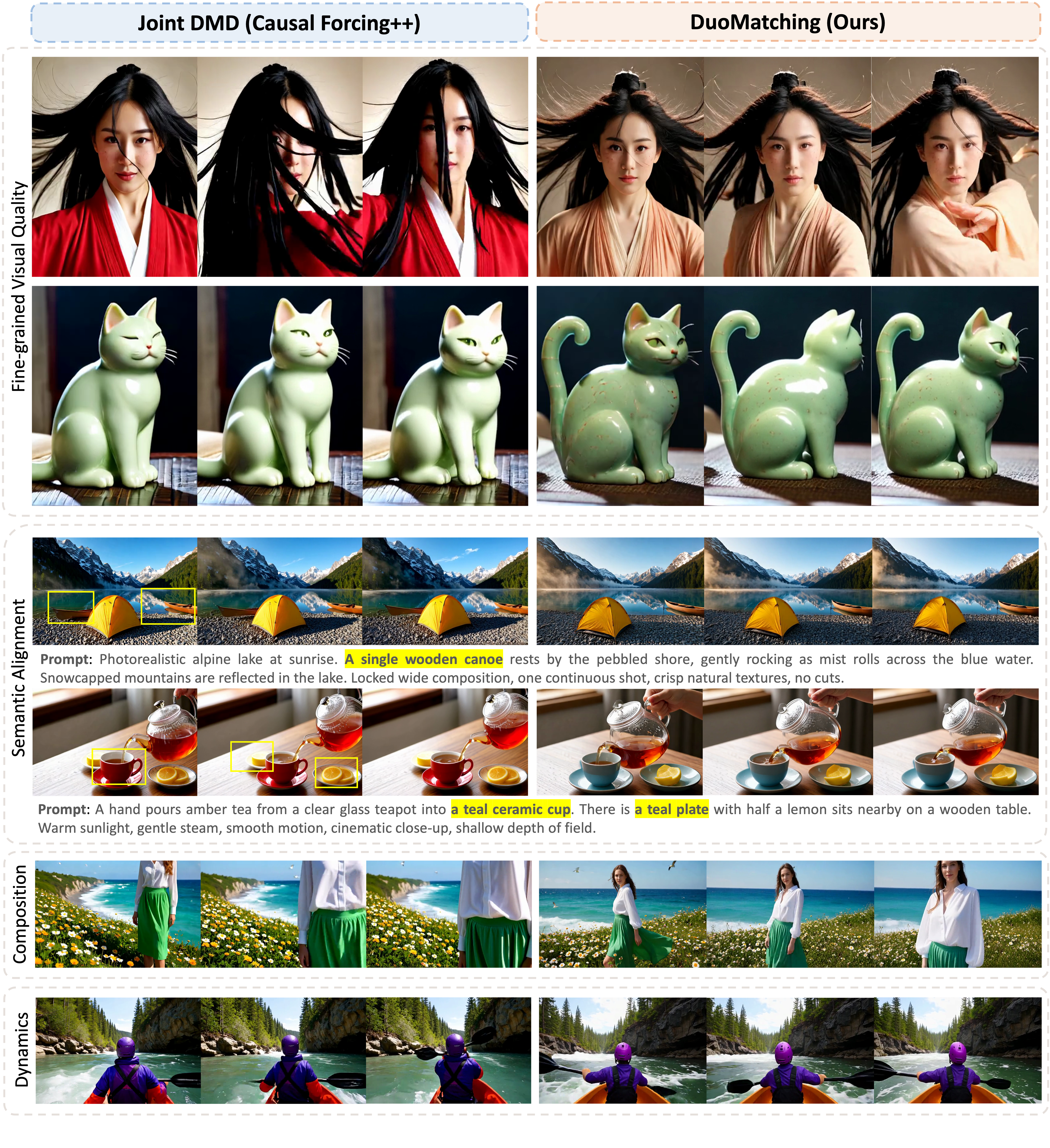}
\caption{Qualitative comparison between \ourmethod and Causal Forcing++ across four aspects: fine-grained visual quality, semantic alignment, composition, and dynamics. \ourmethod improves visual details, prompt adherence, and composition while preserving video dynamics.}
    \label{fig:qualitative_result}
\end{figure}

\myparagraph{Implementation details.} 
For causal models, we initialize the generator from the causal consistency distillation checkpoint from Causal Forcing++~\citep{zhao2026causal} and subsequently optimize it with DMD on VidProM~\citep{wang2024vidprom} for 1,000 training steps. For bidirectional models, following CausVid~\citep{yin2025slow}, we initialize the generator directly from Wan2.1-T2V-1.3B and perform DMD training for 1200 steps on Mixkit~\citep{lin2024open}. 
In both settings, we use the frozen Qwen-Image~\citep{wu2025qwen} as the default image teacher.
The output videos consist of $81$ frames and have a resolution of $H{=}480$, $W{=}832$.
\submethod is trained separately on a curated subset of OpenVid~\citep{nan2025openvid} containing 29,400 videos with high aesthetic quality and substantial motion.
For \ourmethod, we set the marginal DMD loss weight to $\omega=0.4$ and sample $K=4$ temporal latent slices from each video by default. 
All experiments are conducted on eight GPUs, each with 80 GB of memory.

\myparagraph{Baselines.}
For causal generation, we compare our method with autoregressive video generation methods, including Self Forcing~\citep{huang2026self}, Reward Forcing~\citep{lu2026reward},  Causal Forcing++~\citep{zhao2026causal,zhu2026causal} and One-Forcing~\citep{feng2026one}, covering both chunk-wise and frame-wise variants when available. For bidirectional generation, we use the bidirectional DMD variant of CausVid~\citep{yin2025slow} as our primary baseline. 

\myparagraph{Metrics.}
We conduct automatic evaluation on VBench~\citep{huang2024vbench} using its official prompt suite and evaluation protocol. 
To more thoroughly assess video generation quality in fine-grained visual quality, visual composition, and semantic alignment, we additionally curate a test set of 400 challenging prompts featuring rich visual details and complex dynamics. We report the Semantic Score, Aesthetic Quality, Imaging Quality, Dynamic Degree, and Motion Smoothness, together with the aggregate VBench score. 

\subsection{Results}

\myparagraph{Qualitative comparison with baselines.} Figure~\ref{fig:qualitative_result} presents side-by-side comparisons between Causal Forcing++ and \ourmethod. 
Compared with the baseline, \ourmethod produces finer visual details, such as more distinct surface textures and subtle color variations on the ceramic cat.
Benefiting from the semantic alignment capabilities of the image teacher, \ourmethod also better follows prompt specifications, including object counts (a single wooden canoe) and attributes (a teal ceramic cup). 
In addition, \ourmethod exhibits improved composition, framing the subject as intended rather than incorrectly focusing the camera on the waist. 
These improvements in fine-grained visual quality, composition, and semantic alignment are achieved while preserving temporal quality, as illustrated by the smooth and consistent paddling motions in the kayaking example.
Further comparisons with other baselines are provided in Appendix~\ref{sec:more_qualitative}.

\begin{table*}[t]
\centering
\caption{
Quantitative comparison on VBench with Self Forcing~\citep{huang2026self}, One-Forcing~\citep{feng2026one}, Reward Forcing~\citep{lu2026reward} and Causal Forcing++~\citep{zhao2026causal}.
``Full'' denotes bidirectional full-video generation,
while ``Frame'' and ``Chunk'' denote the causal autoregressive
generation unit.
NFE denotes the number of function evaluations per generation unit.
Results should be compared within matched generation modes and
NFE budgets.
Bold indicates the best result within each matched group;
ours are shaded in light gray.
}
\label{tab:main_vbench}
\small
\setlength{\tabcolsep}{3.2pt}
\begin{tabular}{lcccccccc}
    \toprule
    Method
    & Mode
    & NFE
    & Semantic $\uparrow$
    & Aesthetic $\uparrow$
    & Imaging $\uparrow$
    & Dynamic $\uparrow$
    & Smoothness $\uparrow$
    & Total $\uparrow$ \\

    \midrule
    \multicolumn{9}{l}{
        \textit{Bidirectional DMD methods}
    } \\

    CausVid
    & Full & 4
    & 69.23 & 64.15 & 73.00
    & 36.67 & \textbf{98.84} & 80.63 \\

    \rowcolor{oursgray}
    \textbf{Ours}
    & Full & 4
    & \textbf{70.89} & \textbf{64.55} & \textbf{73.77}
    & \textbf{39.17} & 98.82 & \textbf{81.45} \\

    \midrule
    \multicolumn{9}{l}{
        \textit{Causal autoregressive methods}
    } \\

    \addlinespace[2pt]
    \multicolumn{9}{l}{
        \hspace{0.8em}\textit{Frame-wise generation (1 step)}
    } \\

    Causal Forcing++
    & Frame & 1
    & 57.17 & 67.12 & 68.83
    & \textbf{70.56} & 98.49 & 79.82 \\

    One-Forcing
    & Frame & 1
    & 69.25 & 65.08 & 68.28
    & 66.11 & 99.03 & 82.09 \\

    \rowcolor{oursgray}
    \textbf{Ours}
    & Frame & 1
    & \textbf{69.95} & \textbf{68.62} & \textbf{69.42}
    & 69.44 & \textbf{99.07} & \textbf{82.65} \\

    \addlinespace[3pt]
    \multicolumn{9}{l}{
        \hspace{0.8em}\textit{Frame-wise generation (2 steps)}
    } \\

    Causal Forcing++
    & Frame & 2
    & 68.42 & 62.28 & 69.82
    & \textbf{94.44} & 98.20 & 80.67 \\

    \rowcolor{oursgray}
    \textbf{Ours}
    & Frame & 2
    & \textbf{72.84} & \textbf{66.97} & \textbf{73.48}
    & 93.61 & \textbf{98.39} & \textbf{83.53} \\

    \addlinespace[3pt]
    \multicolumn{9}{l}{
        \hspace{0.8em}\textit{Chunk-wise generation (4 steps)}
    } \\

    CausVid
    & Chunk & 4
    & 68.69 & 62.16 & 66.97
    & 55.83 & 98.04 & 80.19 \\

    Self Forcing
    & Chunk & 4
    & 71.41 & 66.83 & 69.66
    & 68.61 & 95.52 & 81.52 \\

    Reward Forcing
    & Chunk & 4
    & 56.23 & 63.62 & 71.72
    & 73.61 & 98.44 & 80.10 \\

    Causal Forcing++
    & Chunk & 4
    & 70.84 & 64.47 & 70.13
    & \textbf{80.56} & 98.14 & 82.79 \\

    \rowcolor{oursgray}
    \textbf{Ours}
    & Chunk & 4
    & \textbf{71.61} & \textbf{66.90} & \textbf{72.41}
    & 76.67 & \textbf{99.06} & \textbf{83.51} \\

    \bottomrule
\end{tabular}
\vspace{-1em}
\end{table*}

\begin{table*}[t]
    \centering
    \caption{
        Human preference study under the same Wan2.1 video teacher.
        Each entry reports the percentage of pairwise judgments favoring our method over the corresponding baseline.
    }
    \label{tab:human_study}
    \small
    \setlength{\tabcolsep}{5pt}

  \begin{tabular}{lcccc}
    \toprule
    Comparison
    & Visual Quality $\uparrow$
    & Semantic Alignment $\uparrow$
    & Temporal \& Motion $\uparrow$
    & Overall $\uparrow$ \\
    \midrule
    Ours vs.\ CausVid
    & 92.17\%& 89.13\%& 51.30\%& 91.30\%\\

    Ours vs.\ Causal Forcing++
    & 81.30\%& 79.57\%& 49.57\%& 80.87\%\\

    Ours vs.\ One-Forcing
    & 93.48\%& 91.74\%& 90.87\%
    & 94.35\%\\

    Ours vs.\ Reward Forcing
    & 83.91\%& 82.61\%
    & 60.87\%& 83.91\%\\
    \bottomrule
\end{tabular}

\vspace{-1em}
\end{table*}

\begin{table}[t]
\centering
\caption{
Ablation of image teachers, comparing Wan2.1-14B~\citep{wan2025wan}, SDXL~\citep{podell2024sdxl}, FLUX.2-4B~\citep{flux-2-2025}, and Qwen-Image~\citep{wu2025qwen} against a baseline without an image teacher.
Teacher HPSv3~\citep{ma2025hpsv3} measures human preference alignment of images generated by each teacher, while the remaining metrics evaluate the resulting video student.
}
\label{tab:image_teacher_ablation}
\small
\setlength{\tabcolsep}{3.5pt}
\begin{tabular}{lccccccc}
    \toprule
    & Teacher & \multicolumn{6}{c}{Video Student} \\
    \cmidrule(lr){2-2}
    \cmidrule(lr){3-8}
    Image Teacher
    & HPSv3 $\uparrow$
    & Semantic $\uparrow$
    & Aesthetic $\uparrow$
    & Imaging $\uparrow$
    & Dynamic $\uparrow$
    & Smoothness $\uparrow$
    & Total $\uparrow$ \\
    \midrule
    None
    & --
    & 68.42
    & 62.28
    & 69.82
    & 94.44
    & 98.20
    & 80.67 \\

    Wan2.1-14B
    & 5.87
    & 68.77
    & 64.16
    & 69.90
    & 94.72
    & 99.05
    & 81.43 \\

    SDXL
    & 6.57
    & 68.50
    & 64.78
    & 70.01
    & 93.89
    & 98.12
    & 81.47 \\

    FLUX.2-4B
    & 8.86
    & 69.06
    & 65.29
    & 71.58
    & 94.17
    & 99.23
    & 82.64 \\

    Qwen-Image
    & 9.05
    & 72.84
    & 66.97
    & 73.48
    & 93.61
    & 98.39
    & 83.53 \\

    \bottomrule
\end{tabular}
\end{table}

\begin{table}[t]
    \centering
\caption{
    Ablation of marginal DMD and latent mapping strategies.
    Direct applies marginal DMD directly to temporally compressed video latent slices.
    Decode-Encode decodes video latents into RGB frames and re-encodes sampled frames with the image VAE, retaining gradients through both operations.
    Peak memory denotes the maximum GPU memory allocated to tensors during a complete training iteration.
    OOM indicates an out-of-memory failure.
}
    \label{tab:latentbridge_ablation}
    \small
    \setlength{\tabcolsep}{4pt}
    \begin{tabular}{ccccccc}
        \toprule
        \multicolumn{2}{c}{Configuration}
        & \multicolumn{4}{c}{Evaluation Metrics}
        & Peak Memory \\
        \cmidrule(lr){1-2}
        \cmidrule(lr){3-6}
        Marginal DMD
        & Latent Mapping
        & Semantic $\uparrow$
        & Imaging $\uparrow$
        & Dynamic $\uparrow$
        & Total $\uparrow$
        & (GB) $\downarrow$ \\
        \midrule
        $\times$
        & ---
        & 68.42 & 69.82 & 94.44 & 80.67
        & 56.49 \\

        $\checkmark$
        & Direct
        & 71.74 & 71.66 & 76.94 & 81.78
        & 59.40 \\

        $\checkmark$
        & Decode-Encode
        & --- & --- & --- & ---
        & OOM \\

        $\checkmark$
        & \submethod
        & 72.84 & 73.48 & 93.61 & 83.53
        & 59.41 \\
        \bottomrule
    \end{tabular}
\end{table}

\myparagraph{Quantitative results.}
The quantitative results in Table~\ref{tab:main_vbench} show \ourmethod's significant advantages over the compared methods within each matched generation mode and NFE budget. 
For bidirectional generation, \ourmethod improves upon CausVid in visual quality, semantic alignment, and video dynamics. 
For causal generation, \ourmethod achieves higher video generation quality across one-, two-, and four-step settings, yielding considerable improvements in visual quality and semantic alignment while largely preserving motion quality.
These results support our joint-marginal formulation, which improves frame-level quality through explicit marginal matching and complementary visual and semantic priors from pretrained image generation models.
All gains are achieved without additional inference computation, preserving the real-time generation capability of Causal Forcing++~\citep{zhao2026causal}.

\myparagraph{Human evaluation.}
We further conduct a human evaluation following a two-alternative forced-choice (2AFC) protocol. 
We recruit 23 participants, each of whom evaluates 40 prompt-matched pairs of videos generated by different methods and indicates their preference in terms of visual quality, semantic alignment, temporal quality, and overall quality.
As shown in Table~\ref{tab:human_study}, \ourmethod achieves overall preference rates above 80\% against every baseline. 
It receives higher preference rates for visual quality and semantic alignment, supporting the effectiveness of explicit marginal distribution matching in improving frame-level quality.
Preference rates for Temporal \& Motion are near or above parity, indicating that temporal quality is largely preserved.

\subsection{Ablation Study}

\myparagraph{Gains from marginal matching and stronger image priors.}
Our ablation suggests that the improvements above stem from both explicit marginal matching and stronger image priors.
As shown in Table~\ref{tab:image_teacher_ablation}, using Wan2.1-14B as the image teacher, without introducing additional priors from an external image generation model, already improves performance across all reported metrics.
We further compare SDXL~\citep{podell2024sdxl}, FLUX.2-4B~\citep{flux-2-2025}, and Qwen-Image~\citep{wu2025qwen}.
Across these three image teachers, higher teacher HPSv3 scores are accompanied by consistent improvements in Semantic, Aesthetic, and Imaging scores.
Thanks to the spatiotemporal decoupling enabled by \submethod, Dynamic Degree and Motion Smoothness remain stable across different image teachers, indicating that teacher capability primarily affects visual quality and semantic alignment with relatively trivial changes in motion dynamics and smoothness.

\myparagraph{Enabling frame-level marginal matching with \submethod.} 
We investigate the role of \submethod in enabling marginal matching. 
Qwen-Image and Wan share the same VAE encoder, allowing us to directly apply marginal DMD to temporally compressed video latent slices as a baseline. 
As shown in Table~\ref{tab:latentbridge_ablation}, such direct application improves semantic alignment and visual quality, but suppresses the dynamic information encoded in the latent representation, resulting in videos with limited motion. 
By mapping temporally compressed video latents to frame-specific image latents, \submethod separates spatial information in video latents and aligns it with the target image latent space.
\submethod yields further gains in visual quality and semantic alignment while largely retaining video dynamics.
While Decode-Encode runs out of memory under our training configuration, \submethod incurs only a small additional memory overhead over the Direct baseline.

\myparagraph{Effective allocation of frame-level supervision with Latent Variation Sampling (LVS).}
We first examine the effect of the number of sampled latent slices $K$, then compare LVS with uniform random and temporally stratified sampling.
As shown in Table~\ref{tab:sampling_ablation}, increasing $K$ from $2$ to $4$ improves the Total score for both uniform sampling and LVS, whereas further increasing $K$ to $8$ reduces both Dynamic and Total scores. This indicates that increasing $K$ raises computational cost without necessarily improving performance, highlighting the importance of an effective supervision allocation strategy.
LVS outperforms uniform sampling across all tested $K$ and temporally stratified sampling at $K=4$.
These results demonstrate that LVS enables more effective frame-level supervision while preserving temporal quality and limiting computational cost.

\begin{table}[H]
    \centering
\caption{Ablation of the number of sampled latent slices $K$ and sampling strategies. Temporally stratified sampling randomly selects one latent slice from each of $K$ segments of equal length. }
    \label{tab:sampling_ablation}
    \small
    \setlength{\tabcolsep}{7pt}
    \renewcommand{\arraystretch}{1.15}

    \begin{tabular}{lccc}
        \toprule
        Sampling strategy
        & $K=2$
        & $K=4$
        & $K=8$ \\
        \cmidrule(lr){2-4}
        & Dynamic $\uparrow$ / Total $\uparrow$
        & Dynamic $\uparrow$ / Total $\uparrow$
        & Dynamic $\uparrow$ / Total $\uparrow$ \\
        \midrule

        Uniform random
        & 93.06 / 80.99& 91.39 / 82.61& 89.44 / 81.04\\

        Temporally stratified
        & -- / -- & 91.67 / 82.68& -- / -- \\

        LVS (Ours)
        & 94.17 / 81.15& 93.61 / 83.53&  92.22 / 82.55\\

        \bottomrule
    \end{tabular}
\end{table}

\section{Conclusion}
We presented \ourmethod, a unified joint-marginal distribution matching framework for few-step video generation. By complementing joint DMD with explicit marginal DMD, our framework strengthens frame-level supervision and transfers visual and semantic priors from image models. \submethod enables marginal matching for temporally compressed video latents, while Latent Variation Sampling reduces redundant supervision. Experiments demonstrate improvements in visual quality, composition, and semantic alignment while largely preserving motion dynamics.

\FloatBarrier
\section*{Acknowledgments}
We thank Shuai Ma for his suggestions on the name \ourmethod and Kunyu Feng and Zhenhua Yang for their helpful suggestions on the writing of this paper.

\bibliographystyle{plainnat}
\bibliography{references}

\clearpage
\appendix

\section{Hyperparameter ablations for \ourmethod.}
We examine the effect of the marginal DMD loss weight $\omega$. As shown in Table~\ref{tab:weight_ablation}, increasing $\omega$ from $0.1$ to $0.4$ improves semantic alignment and imaging quality while maintaining a similar Dynamic score. Further increasing $\omega$ to $0.8$ reduces both metrics and substantially lowers the Dynamic score, suggesting that excessive marginal supervision compromises the balance between visual quality and video dynamics. Among the tested settings, $\omega=0.4$ achieves the highest scores across all reported metrics, and we adopt it as the default.

\begin{table}[htbp]
    \centering
    \caption{
        Ablation of the marginal DMD loss weight $\omega$.
    }
    \label{tab:weight_ablation}
    \small
    \setlength{\tabcolsep}{7pt}

    \begin{tabular}{lccc}
        \toprule
        Metric
        & $\omega=0.1$
        & $\omega=0.4$
        & $\omega=0.8$ \\
        \midrule

        Semantic $\uparrow$
        & 69.60 & 72.84 & 71.49 \\

        Imaging $\uparrow$
        & 70.79 & 73.48 & 71.74 \\

        Dynamic $\uparrow$
        & 93.61 & 93.61 & 87.78 \\

        Total $\uparrow$
        & 81.03 & 83.53 & 81.57 \\

        \bottomrule
    \end{tabular}
\end{table}

\section{Additional Qualitative Comparisons}
\label{sec:more_qualitative}

\begin{figure}[!t]
    \centering
    \includegraphics[width=\textwidth,height=0.70\textheight,keepaspectratio]{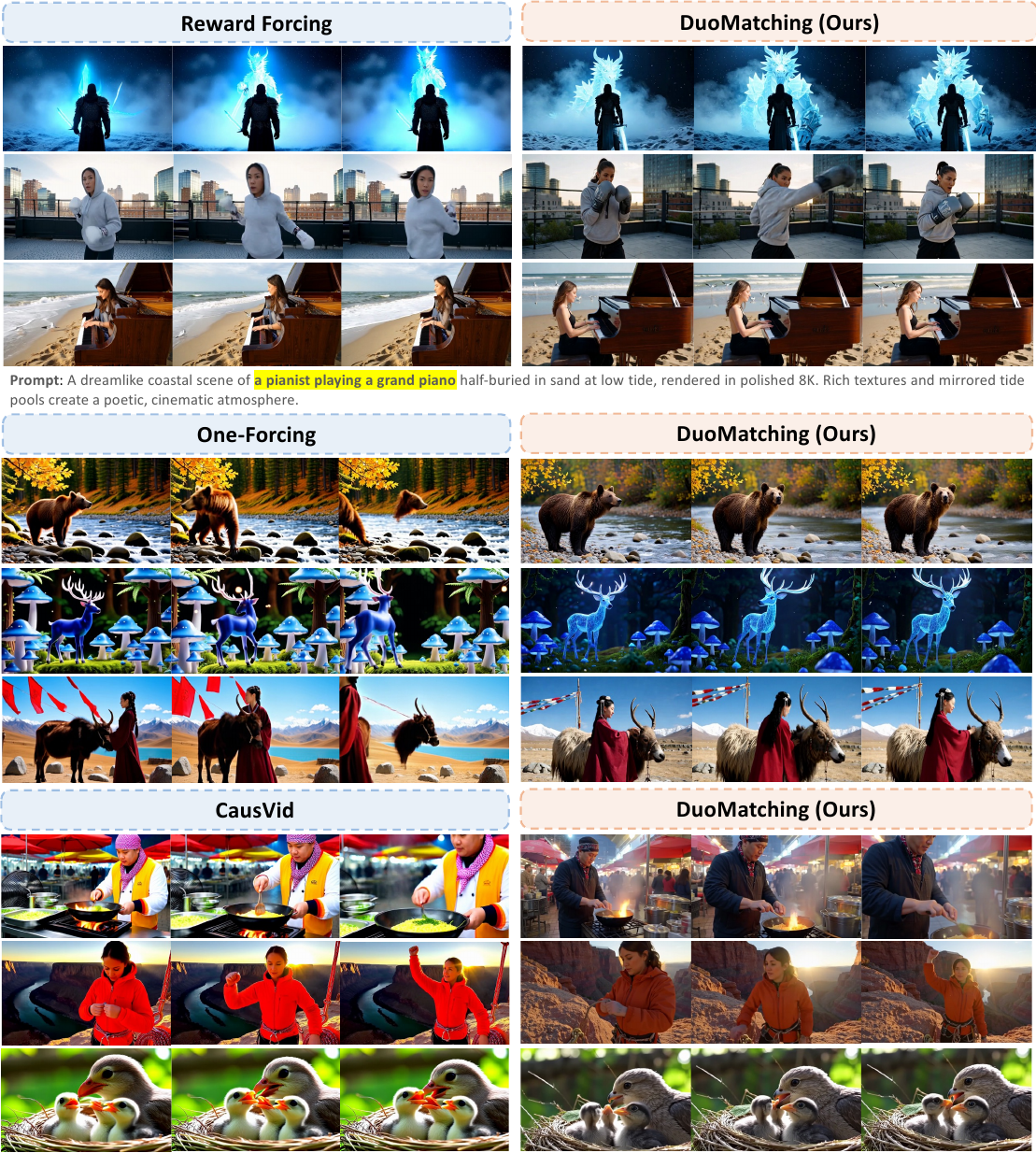}
    \caption{Additional qualitative comparisons of \ourmethod with Reward Forcing, One-Forcing, and CausVid. \ourmethod improves visual details, semantic alignment, and composition while preserving video dynamics.}
    \label{fig:more_qualitative_result}
\end{figure}

Figure~\ref{fig:more_qualitative_result} presents additional qualitative comparisons with Reward Forcing, One-Forcing, and CausVid. Compared with Reward Forcing, \ourmethod renders the ice dragon with more clearly defined structures and finer details, while better preserving the boxer's appearance during rapid motion. In the pianist example, it also produces a more plausible spatial relationship between the pianist, keyboard, and piano, better reflecting the intended interaction. Compared with One-Forcing and CausVid, \ourmethod alleviates overexposure and produces more realistic results, with more natural lighting and richer details in animal fur, feathers, clothing, and surrounding environments. In particular, under the $NFE=1$ setting, \ourmethod substantially improves motion stability over One-Forcing, reducing abrupt changes in subject appearance and position across frames. Together, these examples illustrate improvements in generation quality across different video distillation baselines.

\section{Proof of the Marginal Fitting Comparison}
\label{app:marginal_reference}

Using the notation of the main text, let $P^*$ denote the real video distribution, $P_v$ the video reference, and $P_i$ the image reference. Using the same frame-sampling operator $M$, define $m^*=MP^*$ and $m_v=MP_v$. Let $\mathcal P$ be a feasible family of video distributions and let $\mathcal F=\{MP:P\in\mathcal P\}$. For this idealized comparison, assume matched marginal capacity: $\mathcal F$ is also the feasible family of image distributions. We assess reference fitting using forward KL.

For any $P\in\mathcal P$, construct the joint distributions of a video $x$ and a sampled frame $y$ using the same sampling kernel $M(y\mid x)$. The KL chain rule gives
\begin{equation}
D_{\mathrm{KL}}(P^*\|P)
=
D_{\mathrm{KL}}(m^*\|MP)+C(P),
\end{equation}
where
\begin{equation}
C(P)
=
\mathbb E_{y\sim m^*}
\left[
D_{\mathrm{KL}}
\bigl(P^*(x\mid y)\|P(x\mid y)\bigr)
\right].
\end{equation}
The conditional distributions are defined by this video-and-frame sampling procedure. Thus, joint fitting accounts for both marginal and conditional distribution discrepancies.

In this idealized comparison, assume that $P_i$ and $P_v$ attain the following finite minima:
\begin{equation}
P_i\in\arg\min_{m\in\mathcal F}
D_{\mathrm{KL}}(m^*\|m),
\qquad
P_v\in\arg\min_{P\in\mathcal P}
D_{\mathrm{KL}}(P^*\|P).
\end{equation}
Since $P_i\in\mathcal F$, there exists a video distribution $P^\dagger\in\mathcal P$ whose frame-level marginal satisfies $MP^\dagger=P_i$. We assume that such a distribution can be chosen with $D_{\mathrm{KL}}(P^*\|P^\dagger)<\infty$. This auxiliary distribution allows us to compare joint fitting under $P_v$ with fitting the image reference $P_i$ as the marginal.

The marginal optimality of $P_i$ gives
\begin{equation}
D_{\mathrm{KL}}(m^*\|P_i)
\le
D_{\mathrm{KL}}(m^*\|m_v).
\end{equation}
The joint optimality of $P_v$, together with the KL decomposition, gives
\begin{equation}
D_{\mathrm{KL}}(m^*\|m_v)+C(P_v)
\le
D_{\mathrm{KL}}(m^*\|P_i)+C(P^\dagger).
\end{equation}
Combining these inequalities yields
\begin{equation}
0
\le
D_{\mathrm{KL}}(m^*\|m_v)-D_{\mathrm{KL}}(m^*\|P_i)
\le
C(P^\dagger)-C(P_v).
\end{equation}

Thus, under matched marginal capacity, marginal-only fitting achieves no greater marginal error than joint fitting, and this advantage is bounded by the additional conditional fitting error of $P^\dagger$ relative to $P_v$.

\section{Proof of Joint-Marginal Improvement}
\label{app:joint_marginal_improvement}

Using the same frame-sampling operator $M$, define $m^*=MP^*$ and $m_v=MP_v$. Assume that the image reference has no greater marginal approximation error than the video-teacher marginal, measured by forward KL, and that the two distributions differ:
\begin{equation}
D_{\mathrm{KL}}(m^*\|P_i)
\le D_{\mathrm{KL}}(m^*\|m_v),
\qquad P_i\ne m_v.
\end{equation}
Consider the ideal distribution-space optimum
\begin{equation}
Q_\omega\in\arg\min_Q
\left\{
D_{\mathrm{KL}}(Q\|P_v)
+\omega D_{\mathrm{KL}}(MQ\|P_i)
\right\},
\qquad m_\omega=MQ_\omega,
\end{equation}
where the minimization is over all video distributions. We assume that the reference and optimal distributions have positive densities on common supports and that all displayed KL divergences are finite. We further assume that the optimal density path $Q_\omega$ exists for all sufficiently small $\omega\geq 0$, is right-differentiable at $\omega=0$, and permits interchanging the differentiations below with integration and the sampling operator. We use the same symbols for distributions and their densities. At $\omega=0$, the unique minimizer is $Q_0=P_v$. To establish local improvement, we show that $D_{\mathrm{KL}}(P^*\|Q_\omega)$ has a strictly negative right derivative at $\omega=0$.

Since $M$ is linear, the first variations of the two KL terms with respect to $Q(x)$ are $\log(Q(x)/P_v(x))+1$ and $\mathbb E_{Y\sim M(\cdot\mid x)}[\log((MQ)(Y)/P_i(Y))+1]$, respectively.
\begin{equation*}
\begin{aligned}
(MQ)(y)
&=\int M(y\mid x)Q(x)\,dx,
\qquad
\frac{\delta(MQ)(y)}{\delta Q(x)}=M(y\mid x),\\
\frac{\delta}{\delta Q(x)}
D_{\mathrm{KL}}(MQ\|P_i)
&=\int M(y\mid x)
\left[
\log\frac{(MQ)(y)}{P_i(y)}+1
\right]\,dy.
\end{aligned}
\end{equation*}
\begin{equation*}
\mathcal A_\omega(Q,\lambda)
=
D_{\mathrm{KL}}(Q\|P_v)
+\omega D_{\mathrm{KL}}(MQ\|P_i)
+\lambda\left(\int Q(x)\,dx-1\right).
\end{equation*}
Imposing the normalization constraint and absorbing terms independent of $x$ into a constant gives
\begin{equation*}
\begin{aligned}
0
&=
\log\frac{Q_\omega(x)}{P_v(x)}+1
+\omega\,
\mathbb E_{Y\sim M(\cdot\mid x)}
\left[
\log\frac{m_\omega(Y)}{P_i(Y)}+1
\right]
+\lambda_\omega,\\
c_\omega&=-1-\omega-\lambda_\omega.
\end{aligned}
\end{equation*}
\begin{equation}
\log\frac{Q_\omega(x)}{P_v(x)}
+\omega\,\mathbb E_{Y\sim M(\cdot\mid x)}
\left[\log\frac{m_\omega(Y)}{P_i(Y)}\right]
=c_\omega.
\end{equation}
Define
\begin{equation}
g(x)=\mathbb E_{Y\sim M(\cdot\mid x)}
\left[\log\frac{m_v(Y)}{P_i(Y)}\right].
\end{equation}
Taking the right derivative of the optimality condition at $\omega=0$, with dots denoting these derivatives, yields
\begin{equation}
\begin{aligned}
\dot c_0
&=
\frac{\dot Q_0(x)}{Q_0(x)}
+
\mathbb E_{Y\sim M(\cdot\mid x)}
\left[
\log\frac{m_0(Y)}{P_i(Y)}
\right]\\
&=
\frac{\dot Q_0(x)}{P_v(x)}+g(x).
\end{aligned}
\end{equation}
Differentiating $\int Q_\omega(x)\,dx=1$ gives $\int\dot Q_0(x)\,dx=0$. Therefore, integrating the preceding equality against $P_v$ gives
\begin{equation*}
\underbrace{\int\dot Q_0(x)\,dx}_{=0}
+\int P_v(x)g(x)\,dx
=
\dot c_0
\underbrace{\int P_v(x)\,dx}_{=1}.
\end{equation*}
\begin{equation}
\begin{aligned}
\dot c_0
&=\mathbb E_{X\sim P_v}[g(X)]\\
&=\mathbb E_{X\sim P_v}
\mathbb E_{Y\sim M(\cdot\mid X)}
\left[
\log\frac{m_v(Y)}{P_i(Y)}
\right]\\
&=\mathbb E_{Y\sim m_v}
\left[
\log\frac{m_v(Y)}{P_i(Y)}
\right]\\
&=D_{\mathrm{KL}}(m_v\|P_i).
\end{aligned}
\end{equation}

Let $F(\omega)=D_{\mathrm{KL}}(P^*\|Q_\omega)$. Substituting the expression for $\dot Q_0/P_v$ gives
\begin{equation}
\begin{aligned}
F'_+(0)
&=-\int P^*(x)\frac{\dot Q_0(x)}{P_v(x)}\,dx\\
&=-\mathbb E_{X\sim P^*}
\left[\dot c_0-g(X)\right]\\
&=\mathbb E_{X\sim P^*}[g(X)]
-D_{\mathrm{KL}}(m_v\|P_i)\\
&=\mathbb E_{Y\sim m^*}
\left[
\log\frac{m_v(Y)}{P_i(Y)}
\right]
-D_{\mathrm{KL}}(m_v\|P_i)\\
&=\mathbb E_{Y\sim m^*}
\left[
\log\frac{m^*(Y)}{P_i(Y)}
-\log\frac{m^*(Y)}{m_v(Y)}
\right]
-D_{\mathrm{KL}}(m_v\|P_i)\\
&=D_{\mathrm{KL}}(m^*\|P_i)
-D_{\mathrm{KL}}(m^*\|m_v)
-D_{\mathrm{KL}}(m_v\|P_i)\\
&<0.
\end{aligned}
\end{equation}
The last equality follows by expanding the two forward KL divergences. Their difference is nonpositive by assumption, while $D_{\mathrm{KL}}(m_v\|P_i)>0$ because $P_i\ne m_v$. Hence, $F'_+(0)<0$.

By right differentiability, $F(\omega)=F(0)+\omega F'_+(0)+o(\omega)$ as $\omega\downarrow0$. Since $F'_+(0)<0$ and $Q_0=P_v$, there exists $\omega_0>0$ such that
\begin{equation}
D_{\mathrm{KL}}(P^*\|Q_\omega)
<
D_{\mathrm{KL}}(P^*\|P_v),
\qquad 0<\omega<\omega_0.
\end{equation}
Thus, there exists a positive marginal weight $\omega$ for which the ideal joint-marginal optimum is closer to $P^*$ than $P_v$ in forward KL. 

\section{Efficiency and Reconstruction Quality of \submethod}
\label{sec:latentbridge_details}

\begin{figure}[htbp]
    \centering
    \includegraphics[width=0.8\linewidth]{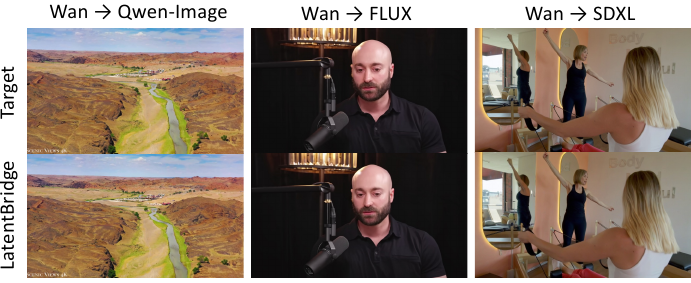}
\caption{
Qualitative evaluation of \submethod across image latent spaces.
Top: target latents obtained by encoding the corresponding RGB frames with each image VAE.
Bottom: reconstructions from latents predicted by \submethod from video latents of Wan2.1.
Both rows use the corresponding image VAE decoder.
}
    \label{fig:latentbridge_reconstruction}
\end{figure}

\submethod provides a lightweight mapping between video and image latent spaces. 
\submethod consists of eight FiLM residual blocks with 256 hidden channels, totaling only 10.75M trainable parameters. 
It is trained on 29,400 videos at 480p resolution with an effective batch size of 32.
Training for 3,000 iterations takes approximately 15 minutes on eight 80-GB GPUs.
Despite its compact size and short training time, \submethod produces reconstructions that closely match the targets in scene structure and visual details, as shown in Figure~\ref{fig:latentbridge_reconstruction}.
The examples across Qwen-Image, FLUX, and SDXL further illustrate its applicability to different image latent spaces.

\section{Transferring Loss Improvement through \submethod}
\label{app:latent_bridge_transfer}
\begin{figure}[htbp]
    \centering
    \includegraphics[width=0.6\textwidth]{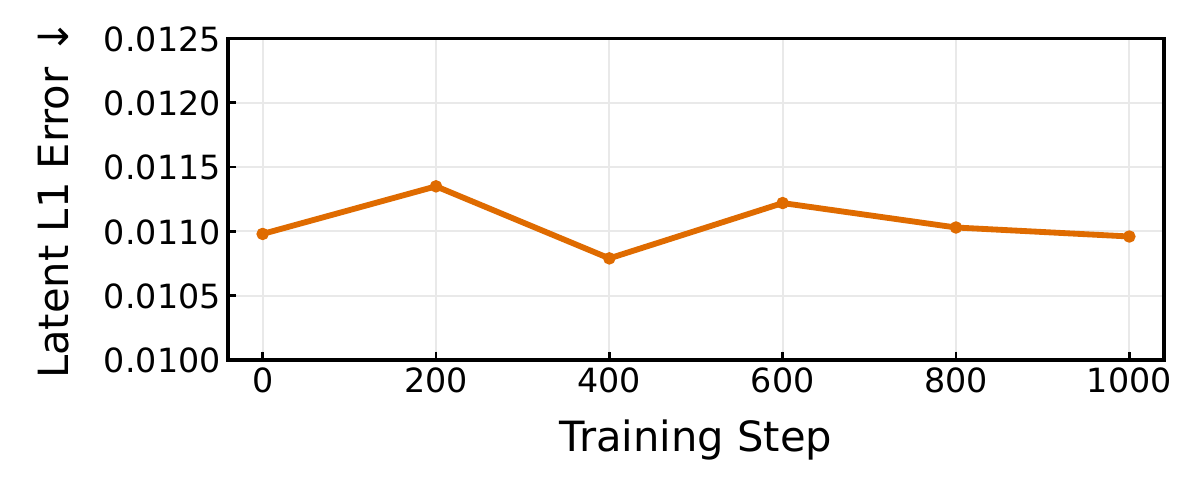}
\caption{Latent reconstruction error of \submethod at different stages of generator training. Given a generated video latent slice $z^l$, with $z^{l-1}$ and the local frame index $i$ as conditioning, \submethod predicts the corresponding image latent. We measure its $\ell_1$ error against the target obtained through Decode-Encode over 100 samples.}
    \label{fig:bridge_latent_l1}
\end{figure}

We establish a sufficient condition under which a loss reduction measured through \submethod also holds for the corresponding decoded video frames. Let $z_\theta=G_\theta(\xi,c)$, where $\xi$ is the generator noise and $c$ is the text condition. For a sampled latent slice $l$ and local frame index $i$, let $t(l,i)$ denote the corresponding RGB frame index. Define the predicted and image-latent representations as
\begin{equation}
u_\theta=B_\phi(z_\theta^{l-1},z_\theta^l,i),
\qquad
v_\theta=E_{\mathrm{img}}\!\left([D_{\mathrm{video}}(z_\theta)]_{t(l,i)}\right),
\end{equation}
where $D_{\mathrm{video}}$ is the video VAE decoder, $E_{\mathrm{img}}$ is the image VAE encoder, and $B_\phi$ is frozen. 
Both representations use the same generated video and sampled frame position. 

Let $\ell(u,c)$ be a fixed scalar frame-level loss, with lower values indicating better performance under this loss. Assume that, on a domain containing the representations considered below, it satisfies
\begin{equation}
|\ell(u,c)-\ell(v,c)|\leq L_\ell\|u-v\|_1
\end{equation}
for a finite constant $L_\ell>0$. Define the two objectives
\begin{equation}
\mathcal J_B(\theta)=\mathbb E[\ell(u_\theta,c)],
\qquad
\mathcal J_R(\theta)=\mathbb E[\ell(v_\theta,c)],
\end{equation}
and assume that these expectations are finite. Consider generator parameters $\theta_0$ and $\theta_1$ before and after an update or a sequence of updates. Suppose the expected mapping errors at these parameters satisfy
\begin{equation}
\mathbb E\|u_{\theta_j}-v_{\theta_j}\|_1\leq\varepsilon_j,
\qquad j\in\{0,1\}.
\end{equation}
The Lipschitz condition then gives, at each endpoint,
\begin{equation}
\begin{aligned}
|\mathcal J_B(\theta_j)-\mathcal J_R(\theta_j)|
&\leq\mathbb E\bigl[|\ell(u_{\theta_j},c)-\ell(v_{\theta_j},c)|\bigr]\\
&\leq L_\ell\mathbb E\|u_{\theta_j}-v_{\theta_j}\|_1
\leq L_\ell\varepsilon_j.
\end{aligned}
\end{equation}

Suppose the objective evaluated through \submethod decreases by $\delta=\mathcal J_B(\theta_0)-\mathcal J_B(\theta_1)>0$. Adding and subtracting the two bridge objectives yields
\begin{equation}
\begin{aligned}
\mathcal J_R(\theta_1)-\mathcal J_R(\theta_0)
&=[\mathcal J_R(\theta_1)-\mathcal J_B(\theta_1)]
  +[\mathcal J_B(\theta_1)-\mathcal J_B(\theta_0)]\\
&\quad+[\mathcal J_B(\theta_0)-\mathcal J_R(\theta_0)]\\
&\leq-\delta+L_\ell(\varepsilon_0+\varepsilon_1).
\end{aligned}
\label{eq:latent_bridge_transfer_bound}
\end{equation}
Consequently, if $\delta>L_\ell(\varepsilon_0+\varepsilon_1)$, the same loss evaluated on image latents encoded from the actual decoded video frames strictly decreases. In particular, when $\varepsilon_0,\varepsilon_1\leq\varepsilon$, a decrease exceeding $2L_\ell\varepsilon$ is sufficient. 

As shown in Figure~\ref{fig:bridge_latent_l1}, the mapping errors remain consistent across training steps. Thus, when these errors remain bounded, a decrease in the fixed frame-level loss through \submethod exceeding $L_\ell(\varepsilon_0+\varepsilon_1)$ guarantees a decrease in the same loss evaluated through Decode--Encode.

\FloatBarrier
\section{Training Procedure}
\label{sec:training_procedure}

Algorithm~\ref{alg:dmd-duo} summarizes the training procedure of \ourmethod.
The generator is optimized with joint DMD and marginal DMD, while both fake estimators are updated on freshly generated samples.
The teachers and \submethod remain frozen, with gradients propagated through \submethod during generator updates.

\begin{algorithm}[htbp]
\caption{Training of \ourmethod}
\label{alg:dmd-duo}
\small
\begin{algorithmic}[1]
\Require Generator $G_\theta$; frozen LatentBridge $B_\phi$
\Require Frozen teachers $\mu_{\mathrm{real}}^{v}$, $\mu_{\mathrm{real}}^{\mathrm{img}}$
\Require Trainable fake estimators $\mu_{\mathrm{fake}}^{v}$, $\mu_{\mathrm{fake}}^{\mathrm{img}}$
\Require Marginal weight $\omega$; sampling budget $K$; update interval $R$; warmup length $T_{\mathrm{warm}}>0$; training iterations $N$

\For{$n=0,\ldots,N-1$}
    \If{$n \bmod R=0$}
        \Statex \hspace{\algorithmicindent}\hspace{\algorithmicindent}%
        \textbf{Generator update}

        \State Sample text condition $c$ and noise $\xi\sim\mathcal{N}(0,I)$
        \State $z\gets G_\theta(\xi,c)$
        \State $\{l_k\}_{k=1}^{K}\gets\operatorname{LVS}(z,K)$

        \For{$k=1,\ldots,K$}
            \State Sample local frame index $i_k$
            \State $u_k\gets B_\phi(z^{l_k-1},z^{l_k},i_k)$
            \Comment{$u_k=u^{l_k,(i_k)}$}
        \EndFor

        \State $U\gets\operatorname{Stack}(u_1,\ldots,u_K)$
        \Comment{Retain gradients through $B_\phi$}
        \State Sample image timestep $\tau$ and $\epsilon\sim\mathcal{N}(0,I)$
        \State $U_\tau\gets\alpha_\tau^{\mathrm{img}}\operatorname{sg}(U)+\sigma_\tau^{\mathrm{img}}\epsilon$

        \State $s_{\mathrm{real}}^{\mathrm{img}}\gets
        \operatorname{Score}(\mu_{\mathrm{real}}^{\mathrm{img}};U_\tau,\tau,c)$
        \Comment{Image teacher; no autograd}

        \State $s_{\mathrm{fake}}^{\mathrm{img}}\gets
        \operatorname{Score}(\mu_{\mathrm{fake}}^{\mathrm{img}};U_\tau,\tau,c)$
        \Comment{Fake estimator; no autograd}

        \State $\Delta s\gets
        s_{\mathrm{fake}}^{\mathrm{img}}-s_{\mathrm{real}}^{\mathrm{img}}$
        \Comment{Score difference}

        \State $g\gets
        \alpha_\tau^{\mathrm{img}}w_{\mathrm{img}}(\tau)\Delta s$
        \Comment{DMD gradient with respect to $U$}

        \State $\widetilde{\mathcal{L}}_{\mathrm{marginal}}\gets
        \frac{1}{2}\operatorname{MSE}\bigl(U,\operatorname{sg}(U-g)\bigr)$
        \Comment{Marginal DMD surrogate loss}

        \State $\mathcal{L}_{\mathrm{joint}}\gets
        \operatorname{JointDMD}(z,c,\mu_{\mathrm{real}}^{v},\mu_{\mathrm{fake}}^{v})$
        \Comment{Joint DMD loss}

        \State $\omega_n\gets\omega\min(1,n/T_{\mathrm{warm}})$
        \State $\widetilde{\mathcal{L}}_{\text{\ourmethod}}\gets
        \mathcal{L}_{\mathrm{joint}}+\omega_n\widetilde{\mathcal{L}}_{\mathrm{marginal}}$
        \Comment{Joint DMD + marginal DMD}
        \State Update $\theta$ using $\nabla_\theta\widetilde{\mathcal{L}}_{\text{\ourmethod}}$
    \EndIf

    \Statex \hspace{\algorithmicindent}%
    \textbf{Fake estimator updates}

    \State Sample fresh text condition $c'$ and noise $\xi'\sim\mathcal{N}(0,I)$
    \State $\bar z\gets\operatorname{sg}(G_\theta(\xi',c'))$
    \State $\{\bar l_k\}_{k=1}^{K}\gets\operatorname{LVS}(\bar z,K)$

    \For{$k=1,\ldots,K$}
        \State Sample local frame index $\bar i_k$
        \State $\bar u_k\gets
        \operatorname{sg}\bigl(B_\phi(\bar z^{\bar l_k-1},\bar z^{\bar l_k},\bar i_k)\bigr)$
    \EndFor

    \State $\bar U\gets\operatorname{Stack}(\bar u_1,\ldots,\bar u_K)$
    \State Sample fresh image timestep $\tau$ and $\epsilon\sim\mathcal{N}(0,I)$
    \State $\bar U_\tau\gets
    \alpha_\tau^{\mathrm{img}}\bar U+\sigma_\tau^{\mathrm{img}}\epsilon$

    \State Update $\mu_{\mathrm{fake}}^{\mathrm{img}}$ with flow-matching target $\epsilon-\bar U$
    \Comment{Fake estimator for marginal DMD}

    \State Update $\mu_{\mathrm{fake}}^{v}$ using the video flow-matching loss on $(\bar z,c')$
    \Comment{Fake estimator for joint DMD}
\EndFor
\end{algorithmic}
\end{algorithm}

\end{document}